\documentclass[twocolumn]{article}
\usepackage{geometry}
\usepackage{authblk}
\usepackage{amsmath, amssymb, amsthm}
\usepackage{pifont}
\usepackage{siunitx}
\usepackage{enumitem}
\usepackage{hyperref}
\usepackage{cite}
\usepackage{xfrac}
\usepackage{siunitx}
\usepackage{comment}
\usepackage{graphicx}
\usepackage{multirow}
\usepackage{tabularx,booktabs}
\usepackage{stfloats}
\usepackage{caption}
\newcounter{supptable}
\renewcommand{\thesupptable}{S\arabic{supptable}}
\usepackage[switch]{lineno}
\newcolumntype{C}{>{\centering\arraybackslash}X} 
\newcolumntype{R}{>{\raggedleft\arraybackslash}X} 
\newcolumntype{L}{>{\raggedright\arraybackslash}X} 
\newcolumntype{P}[1]{>{\centering\arraybackslash}p{#1}}
\newcommand{\cmark}{\ding{51}}
\newcommand{\xmark}{\ding{55}}
\newtheorem*{objective*}{Objective}
\newtheorem{assumption}{Assumption}

\DeclareMathOperator*{\argmin}{argmin}
\makeatletter
\renewcommand\@fnsymbol[1]{\@arabic{#1}}
\makeatother
\title{\fontsize{22pt}{28pt}\selectfont Considering Perspectives for Automated Driving Ethics:\\ Collective Risk in Vehicular Motion Planning}
\author[1,2]{Leon Tolksdorf}
\author[1,3]{Arturo Tejada}
\author[2]{Christian Birkner}
\author[1]{Nathan van de Wouw}
\affil[1]{Department of Dynamics and Control, Eindhoven University of Technology, Eindhoven, The Netherlands, e-mail:
        {\tt\small \{l.t.tolksdorf, a.tejada.ruiz, n.v.d.wouw\}@tue.nl}}
\affil[2]{CARISSMA Institute of Safety in Future Mobility, Technische Hochschule Ingolstadt, Ingolstadt, Germany, e-mail:
        {\tt\small \{leon.tolksdorf, christian.birkner\}@thi.de}}
\affil[3]{TNO, Integrated Vehicle Safety, Helmond, The Netherlands, e-mail:
        {\tt\small arturo.tejadaruiz@tno.nl}}

\date{\vspace{-5ex}}
\begin{document}
% \linenumbers
\twocolumn[
  \maketitle
Recent automated vehicle (AV) motion planning strategies evolve around minimizing risk in road traffic. However, they exclusively consider risk from the AV's perspective and, as such, do not address the ethicality of its decisions for other road users. We argue that this does not reduce the risk of each road user, as risk may be different from the perspective of each road user. Indeed, minimizing the risk from the AV's perspective may not imply that the risk from the perspective of other road users is also being minimized; in fact, it may even increase. To test this hypothesis, we propose an AV motion planning strategy that supports switching risk minimization strategies between all road user perspectives. 
We find that the risk from the perspective of other road users can generally be considered different to the risk from the AV's perspective. Taking a collective risk perspective, i.e., balancing the risks of all road users, we observe an AV that minimizes overall traffic risk the best, while putting itself at slightly higher risk for the benefit of others, which is consistent with human driving behavior.
In addition, adopting a collective risk minimization strategy can also be beneficial to the AV's travel efficiency by acting assertively when other road users maintain a low risk estimate of the AV. Yet, the AV drives conservatively when its planned actions are less predictable to other road users, i.e., associated with high risk. We argue that such behavior is a form of self-reflection and a natural prerequisite for socially acceptable AV behavior.
We conclude that to facilitate ethicality in road traffic that includes AVs, the risk-perspective of each road user must be considered in the decision-making of AVs.
\vspace{1em}
]
Designing automated vehicle (AV) driving behavior requires balancing potentially conflicting interests such as safety, travel efficiency, and adherence to driving ethics. Currently, the responsibility for balancing and defining these aspects primarily falls on the AV designer \cite{bin2022should, sutfeld2025automated}. While safety and travel efficiency may be traded off for each other, i.e., going somewhere faster may be less safe in terms of, e.g., distance to other vehicles, driving ethics is a more subtle optimization criterion as it is challenging to objectively define and, consequently, to measure and control it. \\
We define adherence to driving ethics as considering the costs and benefits of the AV's actions for \textit{all} road users (e.g., a reduction of safety versus an increase in travel efficiency).\\
Existing research has mainly considered these costs and benefits from the AV's perspective. Generating AV behavior by trading off the AV's safety with travel efficiency has given rise to risk-based automated driving strategies. Here, risk is understood as a quantity describing potential harm to the AV given the fact that the AV has limited knowledge, i.e., uncertainty, about current and future states of its surroundings (see, e.g., \cite{Tolksdorf_risk_2025} given a motion planning context, or~\cite{gyllenhammar2025road, salem2024risk} for a safety standardization viewpoint on risk). Risk measures following this definition have been shown to correlate with perceived risk by humans \cite{kolekar2020human, he2024new, krugel2025international}, where \cite{kolekar2020human} has also shown that constraining risk in a static environment (i.e., without other dynamic actors) led to human-comparable vehicle behavior. Trading off risk with progress towards a goal is then referred to as risk-based driving, constituting an AV behavior model. Here, minimizing or constraining risk is established to generate safe AV behavior, see, e.g., \cite{mustafa2024racp,tolksdorf2023risk, nyberg2021risk, schwarting2017safe, ploeg2024risk}.\\
While safe to the AV, the generated AV behavior may not adhere to driving ethics as it could disregard other road users' safety by not taking their perspective into account, which, however, humans do~\cite{krugel2024risk}.
For example, consider a scenario in which an AV is overtaking another slower vehicle from behind, moving through the other vehicle's blind-spot. While cutting in front of the slower vehicle may be objectively safe and, hence, a low-risk maneuver from the AV's perspective, the passengers of the slower vehicle may not feel safe at all, thus perceiving significant risk from their perspective. Existing motion planning literature does not take such different risk perspectives into account, as most reported approaches only use the knowledge on the risk from the AV's perspective or assumes symmetric risk, i.e., the risk being the same from each road user's perspective. \\
Evidently, the capability of AVs to redistribute risks among road users, by its own actions, has clear ethical implications~\cite{krugel2025international}, which, consequently, has led to regulatory attention~\cite{ethicscommission2017, ethicsEU}. Therefore, the work of \cite{geisslinger2023ethical} proposes a motion planning algorithm that minimizes risks along four ethical principles, arguing that their algorithm causes a fair distribution of risk among all road users, and hence, facilitates ethicality. 
That paper gained attention but also sparked a debate,~\cite{kirchmair2023taking}, commenting on the minimization of risks along established ethical principles, that some principles have been taken out of context and may, in certain scenarios, directly oppose each other. \\
Apart from ethical principles guiding the minimization of risks, in \cite{geisslinger2023ethical}, only estimated risks from the AV's perspective are considered. As a result, it is ambiguous whether the proposed planner facilitates ethicality, given the fact that risk may be asymmetric between pairs of road users. In fact, all of the previously discussed risk measures are constituted of two parts, where one part captures the uncertainty in the current and future states of other road users and the other part estimates the severity of colliding with that road user. We argue that asymmetries can arise in both parts. Firstly, a particular collision event can induce a much higher severity for one road user than for another (think of a car-to-truck collision for example). Secondly, also the uncertainty about each other's future motion may be asymmetric, depending on, e.g., the observation history (see blind-spot example above) and the available sensor suite. In~\cite{geisslinger2023ethical}, none of these asymmetries are accounted for and, additionally, their planner's vehicle driving behavior is not explicitly prescribed; for instance, whether or not the AV meets any travel efficiency criteria is left open.\\
The current paper challenges the concept of symmetric risk and explores whether it is beneficial to consider the risk from each road user's perspective within an AV's motion planning algorithm to facilitate ethicality in road traffic. The proposed methodology considers an asymmetric risk function, i.e., the risk may be different from each road user's perspective, where the perspective-based risk is estimated only based on the perception information available to the AV.  
To generate the behavior of an AV, we formulate an optimization problem where risk minimization and travel efficiency maximization are embedded in a joint cost. We solve that optimization problem with a stochastic model predictive controller (SMPC), given the stochastic nature of the risk concept. 
Herewith, we investigate AV behavior that balances collective risk (i.e., considering the risk of the perspective of each road user) with travel efficiency of the AV and compare the resulting behavior to a planning approach based on either an egoistic risk perspective (i.e., only considering the AV's perspective) or on an altruistic risk perspective (i.e., only considering the risk from the other road users' perspectives).\\ 
On the basis of the results of large-scale simulations on real-world and dedicated scenarios, we observe that considering the perspectives of all road users has a significant impact on risk distribution and reduces the overall risk in the driving scene. Here, we find that if the AV allocates between \qty{7.5}{\%} to \qty{8.7}{\%} more risk to itself, it can decrease the risk of each road user by \qty{-8.4}{\%} to \qty{-22.3}{\%}.  Furthermore, the AV acts assertively when road users maintain a good perception of the AV, which is beneficial to the AV's travel efficiency. Yet, the AV drives conservatively when its planned actions are uncertain, i.e., less predictable, to other road users. Such behavior is, to our judgment, a form of self-reflection and a natural prerequisite for safe and socially acceptable AV behavior. Hence, perspective-based risk aids ethically in road traffic.\\

\section*{Results}
To evaluate how the different risk perspectives affect the AV's behavior and overall traffic risk, we perform a quantitative analysis. Hereto, a large-scale simulation study on real-world and hand-crafted scenarios, retrieved from the CommonRoad~\cite{althoff2017commonroad} dataset, is performed.
As we are interested in the specific behavior of the AV (also called the ego vehicle), we select three scenario clusters (\texttt{ZAM\_Tjunction}, \texttt{ZAM\_Zip}, and \texttt{USA\_US101}, respectively) that cover a wide range of maneuvers such as merging, taking a turn, lane-following, etc. Each scenario cluster features roughly similar scenes (i.e., intersection, merging, or highway); however, it involves different initial conditions and maneuvers of the road users and other varying features. Note that we will denote all road users that are not the ego vehicle as objects.
In total, we simulate 683 scenarios across all three scenario clusters. The details of the simulation study and scenario clusters are provided in Supplementary Note 2.
\subsection*{Assumptions and Simulation Setup}
\begin{figure*}[t!]
\begin{center}
\includegraphics[width=16.8cm]{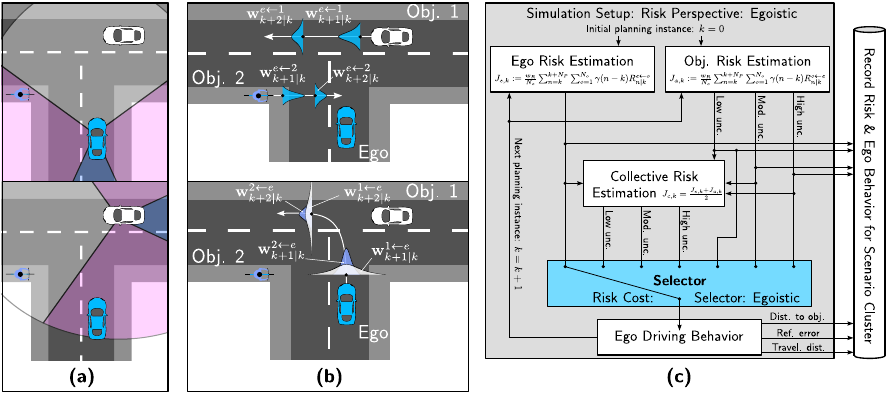}  % The printed column width is 8.4 cm.
\caption{\textbf{(a)} Schematic of a typical traffic scene featuring asymmetric risk. \textit{First}, the aspect of asymmetric uncertainty is depicted by the colored areas. Upper subfigure: the blue vehicle has the white vehicle well in its sight (denoted by its white/grey perception cone); however, it has limited perception (magenta colored cone) of the cyclist, as the cyclist approaches the blue vehicle from its side. Vice-versa in the lower subfigure: the white vehicle has the cyclist well in its forward sight; however, the blue vehicle is in its blind-spot. The cyclist's perception is not depicted, though, clearly, it has both vehicles well in its sight. \textit{Second}, the asymmetric severity can also be deducted. Here, the blue vehicle could collide with its front into the white vehicle's driver-side door. Such a collision is typically associated with more harm to the white vehicle's passengers than to the passengers of the blue vehicle. Likewise, if the cyclist collides with the blue vehicle, the consequences of the collision would be much more severe for the cyclist than for the blue vehicle's passengers.
\textbf{(b)} Example for the assumptions: In the upper subfigure, the ego predicts the uncertain trajectories of two objects, displayed for two different time instances. In the lower subfigure, the ego is estimating the uncertainty both objects have about the planned trajectory of the ego vehicle. The wider distributions for object $1$ indicate that the ego assess that object $1$ is more uncertain about the ego's trajectory than object $2$.
\textbf{(c)} Simulation setup and recording of evaluation metrics for a given scenario with an egoistic risk perspective. The user sets the selector, and a scenario is simulated with that selector setting. While only one risk cost is selected to generate ego behavior, all other risk costs are still recorded (see also Figure~\ref{fig:avg_risk_zip}). Note that the ego's behavior is unaffected by the objects' risks exclusively for the egoistic risk perspective. 
}
\label{fig:assum_sim_workflow}
\end{center}
\end{figure*}
We argue that the risk is different from the perspective of each road user. Therefore, $R_k^{o\gets e}\in \mathbb{R}_{\geq 0}$ denotes the risk of collision of the ego vehicle with object $o \in \mathbb{N}_{>0}$ from the object's perspective, at time instance $k$. Vice-versa, we denote by $R_k^{e\gets o}\in \mathbb{R}_{\geq 0}$ the risk of collision of the ego vehicle with object $o$ from the ego vehicle’s perspective. Risks related to collision events between objects are not considered here.\\ We define risk as the expected collision severity, thus capturing the uncertainty in the estimation of each road user's state and the severity of collision for a combination of those states. A typical driving scene featuring perspective-based risk is depicted in Figure~\ref{fig:assum_sim_workflow}(a).\\
We assume that the ego vehicle predicts the motion of each road user at each time step within a prediction horizon. This prediction is assumed to be stochastic, where the parameterization of the uncertainty is captured in a vector $\mathbf{w}_{n|k}^{e\gets o}$. Here, the subscript $n|k$ denotes a predicted parameterization at time instance $n$ given information available at time $k$, with $n\geq k$. \\
Furthermore, the ego vehicle requires an uncertainty model $\mathbf{w}_{n|k}^{o\gets e}$ for the prediction of its trajectory from the objects' perspectives. Therefore, we assume that the mean of that prediction coincides with the ego's planned trajectory. However, the uncertainty around the mean varies in time and is dependent on the ego's planned maneuver. Such an assumption can be interpreted as the capacity for self-reflection of the ego vehicle in terms of assessing how predictable an action of itself would be to another road user. Without that assumption, the objects' predictions about the ego's states would not affect the ego's motion plan, and hence, the ego would not be enabled to act on the risk perspective of other road users\footnote{Unless there is, e.g., communication or a priori knowledge, which we explicitly do not assume.}. These assumptions are reflected in Figure~\ref{fig:assum_sim_workflow}(b), and additional information is provided in Supplementary Note 1.
In our study, but without loss of methodological generality, we assume that the predictions are Gaussian distributed\footnote{Gaussian distributions are often used to model uncertainties in kinematic variables, see, e.g.,~\cite{dixit2019trajectory, volz2015stochastic, patil2012estimating, ploeg2022long, altendorfer2021new, du2011probabilistic}.}, and hence, $\mathbf{w}_{n|k}^{e\gets o}, \mathbf{w}_{n|k}^{o\gets e}$ contain means and standard deviations of states. \\
As our motion planning strategy plans ego vehicle trajectories, i.e., a sequence of ego vehicle states over a planning horizon of length $N_P \in \mathbb{N}_{>0}$, a method to compose the risks $R_{n|k}^{e\gets o}$, $ R_{n|k}^{o\gets e}$ over a planning horizon, where $ n\in [k, k+1,...,k+N_p]$ for all $N_o\in \mathbb{N}_{>0}$ objects, into a scalar \textit{risk cost} is needed. risk costs are then used as an objective that the motion planning strategy minimizes. To calculate risk costs, we discount the risks $R_{n|k}^{e\gets o}$, $R_{n|k}^{o\gets e}$ over the length of the prediction horizon $N_P$ with the discount function $\gamma$, which can be designed by the user. 
The following three risk costs are defined as:
\begin{equation}\label{eq:risk_costs}
    \begin{split}
        &\text{egoistic risk cost } J_{e,k} := \frac{w_R}{N_o}\sum_{n=k}^{k+N_P}\sum_{o = 1}^{N_o} \gamma(n-k)R_{n|k}^{e \gets o},\\
        &\text{altruistic risk cost }J_{a,k} := \frac{w_R}{N_o}\sum_{n=k}^{k+N_P}\sum_{o = 1}^{N_o} \gamma(n-k)R_{n|k}^{o \gets e},\\
        &\text{collective risk cost }J_{c,k} := \frac{J_{e,k} + J_{a,k}}{2},
    \end{split}
\end{equation}
at time step $k$, where $w_R \in \mathbb{R}_{>0}$ is a weight to balance the risk cost with other costs.
Note that the collective risk cost $J_{c,k}$ does not need to be (independently) computed, as it can always be derived from the egoistic and altruistic risk costs. Furthermore, we will only report the egoistic risk cost $J_{e,k}$, altruistic risk cost $J_{a,k}$, and collective risk cost $J_{c,k}$ throughout the following sections, and not the individual risks $R^{e\gets o}_{n|k}$, and $R^{o\gets e}_{n|k}$, as the risk costs are used by the planner to generate behavior and they are appropriately normalized and weighted. \\
As models predicting the ego's states from the perspective of an object are in their infancy in the literature, we manually assign different uncertainty levels to the object's predictions of the ego to test a broader spectrum of possible object risk perspectives. Here, we test three uncertainty levels, i.e., low, moderate, and high uncertainty. Hence, each scenario within a scenario cluster is simulated seven times to cover the three risk perspectives and object uncertainty modes, the latter of which only apply in the altruistic and collective mode  (namely, in the egoistic mode, the uncertainty that the objects have about the ego does not play a role in the motion planning strategy to begin with). 
Note that, even though the ego vehicle is driving with a given perspective and uncertainty setting in a simulation, we still record the occurring risk costs from all other perspectives. Yet, only the selected risk cost is used to generate (optimized) ego vehicle driving behavior. The risk-based motion planning procedure is illustrated in Figure~\ref{fig:assum_sim_workflow}(c), and further details of the evaluation metrics are given in the section Methods. 
\subsection*{Exemplary Results}
\begin{figure*}[!b]
\begin{center}
\includegraphics[width=16.8cm]{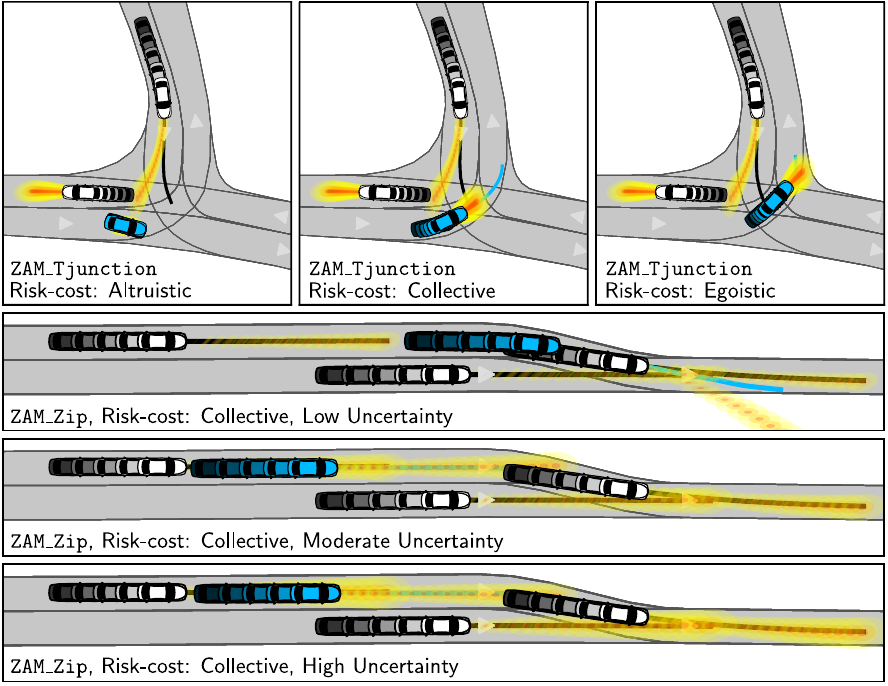}  % The printed column width is 8.4 cm.
\caption{Examples for different risk perspectives (upper subfigures, scenario \texttt{ZAM\_Tjunction-1\_25\_T-1}) and uncertainty settings given a collective risk perspective (lower subfigures, scenario \texttt{ZAM\_Zip-1\_49\_T-1}).
The blue vehicle represents the ego vehicle. The fading colors of vehicles denote a time difference of \qty{0.3}{s} in \texttt{ZAM\_Tjunction} and \qty{0.2}{s} in \texttt{ZAM\_Zip}. The predictions for a prediction horizon of \qty{2}{s} at the presented time step are highlighted by the yellow-to-red ellipses, where a darker color denotes a higher probability. Lastly, the black trajectories of each object denote the respective true trajectory for the prediction horizon, and the blue trajectory shows the planned ego trajectory for the prediction horizon.}
\label{fig:progress}
\end{center}
\end{figure*}

Before performing a large-scale, quantitative analysis, we present selected ego behaviors for the different risk perspectives and different object uncertainties about the ego. A typical example of the ego's behavior for different risk perspectives can be seen in the upper subfigures in Figure \ref{fig:progress}. In the top three figures, we compare all risk perspectives, given moderate object uncertainty about the ego, at a T-junction scenario. Here, the ego is supposed to take a left turn with traffic from its left and oncoming directions. Note that the objects are non-interactive, i.e., the objects' motions are unaffected by the ego's motion. It can be observed that in the altruistic (left) and collective (center) risk perspectives, the ego stops to let an object vehicle pass before taking the turn at the T-junction. However, in the egoistic case (right), the ego vehicle takes the turn before an object enters the T-junction. Additionally, in the collective and egoistic cases, the ego is planning its trajectory so that it sways to the right where the oncoming vehicle is passing on the neighboring lane. \\
Another example is given in the lower subfigures in Figure \ref{fig:progress}, where in all three subfigures the ego vehicle is planning with the collective risk perspective, but the object's uncertainty about the ego's state (configuration and velocity) is altered between low, moderate, and high. In this scenario, the ego is approaching a merging point where it needs to merge between the object vehicles. In the case of low uncertainty about the ego motion, leading to a lower risk perceived by the objects, the ego accelerates and 
merges into a narrow gap between the first two object vehicles. However, for the moderate and high uncertainty cases, leading to higher risks as perceived by the objects, the ego instead decelerates and merges behind the second object vehicle into a larger gap.\\
Both examples illustrate how the ego's behavior changes among different risk perspectives (upper subfigures in Figure~\ref{fig:progress}) and by considering different uncertainties in the objects' perspectives on the ego's motion (lower subfigures in Figure~\ref{fig:progress}). This shows that the asymmetries in the risk perspectives and the way risk is defined influences the ego vehicle behavior and, hence, also the realized risk for all road users and the travel efficiency for the ego vehicle.
\subsection*{Quantitative Results}
The main objective of this paper is to investigate how considering the risk perspectives of other road users in the motion planning strategy of an AV affects, firstly, the risk distribution over all road users and, secondly, the travel efficiency of the AV. The results for dedicated scenarios were discussed above; here, we will present quantitative results for a large-scale simulation campaign. In the following, we vary the risk perspectives and uncertainty settings and present the results for the overall traffic risk, how the risk is distributed, the average risk per scenario cluster, and, lastly, what driving behavior the ego vehicle displays.

\begin{figure*}
\begin{center}
\includegraphics[width=17cm]{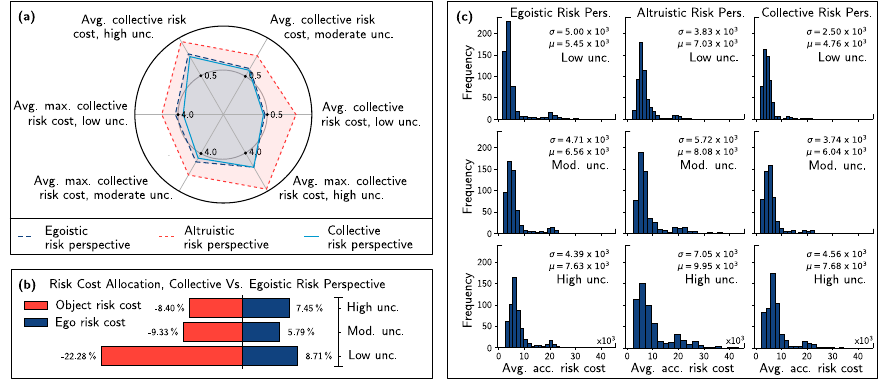}  % The printed column width is 8.4 cm.
\caption{\textbf{(a)} Collective average risk cost for each risk mode (normalized to different scales for visibility). \\ \textbf{(b)} Differences in risk cost allocation between the collective and egoistic perspectives. Negative values indicate a reduction. 
\textbf{(c)} Histograms of average accumulated collective risk cost at the last time step for \texttt{ZAM\_Tjunction}, where, upper row: low uncertainty, center row: moderate uncertainty, bottom row: high uncertainty. Each plot is scaled to contain 18 bins. Note that as we report the average accumulated collective risk cost for each risk-perspective, we obtain nine cases since the uncertainty level always affects the collective risk cost.}
\label{fig:total_risk_histograms}
\end{center}
\end{figure*}

\begin{figure*}[t!]
\begin{center}
\includegraphics[width=16.8cm]{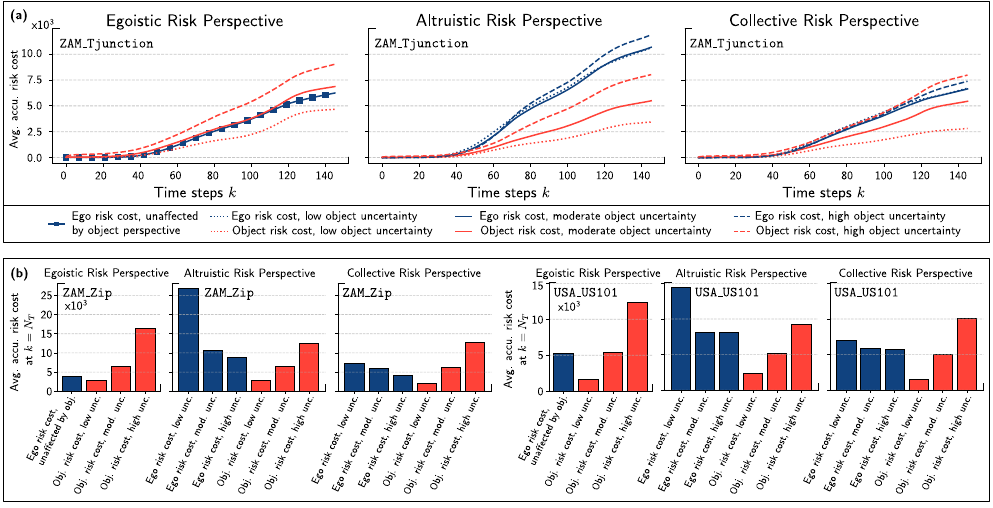}  % The printed column width is 8.4 cm.
\caption{Average accumulated risk costs across scenario clusters, risk perspectives, and uncertainty levels. \textbf{(a)} Average accumulated risk costs over all simulation time steps $k$ for the scenario \texttt{ZAM\_Tjunction}. \textbf{(b)} Average accumulated risk costs for the final time step $k=N_T$ for the scenarios \texttt{ZAM\_Zip} (left) and \texttt{USA\_US101} (right).}
\label{fig:avg_risk_zip}
\end{center}
\end{figure*}

\paragraph{Total Collective Risk Cost}
To investigate the impact of different risk perspectives and uncertainty levels on overall traffic risk, we report the total incurred risk cost. Therefore, the total weighted average collective risk cost and total weighted average maximum collective risk cost (see Evaluation Metrics in Methods) are highlighted in Figure \ref{fig:total_risk_histograms}(a). Here, the altruistic risk perspective performs the worst, as it allocates significant risk to the ego vehicle (see also Figure~\ref{fig:avg_risk_zip}). The collective risk perspective performs equal or better as the egoistic risk perspective in all categories, where the greatest difference is found for low uncertainties in the objects' perspectives. 
\paragraph{Risk Cost Allocation}
Figure~\ref{fig:total_risk_histograms}(b) reports the relative total weighted average accumulated risk cost allocation between the collective and egoistic risk perspectives. It is found that given the collective perspective, the risk cost per object is reduced between \qty{-8.4}{\%} to \qty{-22.28}{\%} when compared to the egoistic perspective. Yet, for the ego, the risk cost is only increased by \qty{7.45}{\%} to \qty{8.71}{\%}. Hence, by the ego vehicle accepting a small increase in risk cost, the risk cost for all other road users can be significantly decreased.

\paragraph{Risk Cost Distribution}
Figure \ref{fig:total_risk_histograms}(c) reports histograms of the average accumulated collective risk costs for scenario cluster \texttt{ZAM\_Tjunction} for each risk perspective and uncertainty level. We choose to highlight scenario cluster \texttt{ZAM\_Tjunction}, as this scenario cluster contains 547 scenarios, which is the largest number of all scenario clusters. In Figure \ref{fig:total_risk_histograms}(c), the spread of the average accumulated collective risk cost is the lowest given the collective risk perspective; only in the high uncertainty case is the spread similar to the egoistic risk perspective. For the altruistic and collective risk perspectives, with increasing uncertainty (center and bottom row), the average accumulated collective risk cost spreads out, indicating that the level of uncertainty of the objects' perspectives about the ego influences the scenario’s outcome. Note that a lower standard deviation is desirable, as such indicates more consistent outcomes from scenario to scenario. The standard deviations of the average accumulated collective risk cost for all scenario clusters are reported in Supplementary Table 1. It shows that the risk spread for low and moderate uncertainty is best for the egoistic perspective; however, closely followed by the collective risk perspective, which, for the high uncertainty cases, outperforms the egoistic perspective. These results indicate that the egoistic and collective risk perspectives yield similar consistency in scenario outcomes.

\paragraph{Average Accumulated risk cost}
The average accumulated risk cost (see Evaluation Metrics in Methods) per scenario cluster is presented in Figure \ref{fig:avg_risk_zip}, for the considered scenario clusters and the three risk perspectives. Note that we only vary the uncertainty level in the objects' perspectives on the ego's motion; thus, we have one uncertainty level from the ego's perspective on the objects' motions. Moreover, for the egoistic risk perspective, the risk cost is unaffected by the object risk cost, and, hence, the ego driving behavior is irrespective of the objects' risks (see also Figure~\ref{fig:assum_sim_workflow}(c)).\\
In the altruistic and collective risk perspectives, the ego's behavior changes with different uncertainty settings of the objects' perspectives of the ego motion (consider different selector settings in Figure~\ref{fig:assum_sim_workflow}(c)). As a result, the egoistic risk cost also differs, as indicated by the dotted and dashed blue graphs (\texttt{ZAM\_Tjunction}) and additional bars (\texttt{ZAM\_Zip} and \texttt{USA\_US101}) for the altruistic and collective risk perspectives.\\
For the scenario cluster \texttt{ZAM\_Tjunction}, we depict the time evolutions of each accumulated risk cost. Here, the ego risk cost is most effectively minimized by the egoistic perspective, where the object's risk costs are minimized to the lowest values for the altruistic perspective. However, in both the egoistic and altruistic perspectives, the perspective that is \textit{not} considered by the planner is put at significantly more risk. It shows that the collective risk perspective performs the best, as every risk cost is nearly as well minimized as for the individual perspective, however, without putting the counterpart at significant additional risk.\\
The results for the scenario clusters \texttt{ZAM\_Zip} and \texttt{USA\_US101} are consistent with the results for \texttt{ZAM\_Tjunction}, where the collective risk minimization balances risks the best while not putting any particular perspective at significant additional risk compared to the individual risk perspective (see Figure~\ref{fig:avg_risk_zip}(b)).\\
Across all scenario clusters, we observe that an altruistic risk perspective is performing unsatisfactorily, as it puts the ego at significant risk, while not minimizing the objects' risk costs consistently to considerably lower values as the collective perspective.\\
Summarizing, we find that the objects' risks vary significantly based on the uncertainty and scenario constellation, indicating that the perspective of the object is an important factor to consider and, indeed, the risk is asymmetric between road users. Furthermore, the collective risk perspective yields the best balance between ego and objects' risks in all scenario clusters.

\paragraph{Ego Driving Behavior}
We also evaluate the ego's driving behavior, given the recorded driving behavior metrics (see Evaluation Metrics in Methods), which we average over all scenarios within a scenario cluster. For the distance to objects, we only report the distance to the closest object.\\
The results are reported in Table \ref{tab:results_scenario_cluster}, where the best values are highlighted in bold font. We consider the lowest reference error\footnote{That is, the L2-norm of the difference between the ego's configuration and velocity with reference values. The reference values are given by following a reference path leading to the goal destination with a reference velocity, see also section Methods.}, the largest average distance to other objects, the largest average traveled distance, and the largest average distance to other objects as best \\
For the scenario clusters \texttt{ZAM\_Zip} and \texttt{USA\_US101}, the best values are associated with the altruistic or collective risk perspectives, where the egoistic perspective leads to the most efficient driving in the \texttt{ZAM\_TJunction} scenario cluster. These differences are likely attributed to the nature of each scenario cluster, where in the \texttt{ZAM\_TJunction} scenario cluster, the ego vehicle tends to take the turn in front of the approaching object. Contrarily, the ego vehicle with the collective and altruistic risk-perspective tends to let the object pass through the intersection before taking the turn. Such can also be observed in the upper subfigures in Figure~\ref{fig:progress}.\\
The general tendency shows that if it is known to the ego that the object's perspectives are associated with low uncertainty towards the ego, the ego drives closer to its references. In this sense, for the ego to consider the object's perspectives has a positive effect on its own travel efficiency. Such is caused by the effect that lower uncertainty causes the spatial risk distribution to be closer to the object vehicles, and, hence, the ego is typically left with more space to navigate. Conversely, the high uncertainty associated with the objects' perspectives can cause the opposite effect, which, however, is desired. Consider, e.g., the blind-spot example from the introduction, where it is not desired that the AV cuts-in in front of another road user, surprising that road user from its blind-spot. 
\begin{table*}[!b]
    \centering
    \caption{Ego behavior for each scenario cluster, an risk setting. Best values are highlighted in bold font.}
    \label{tab:results_scenario_cluster}
    \begin{tabularx}{\linewidth}{RC | C | C | C | C}
        \toprule
        Risk perspective&Uncertainty Level & Avg. Max. Ref. Error & Avg. Acc. Ref. Error & Avg. Travel. Dist. & Avg. Dist. to Closest Obj. \\
        \toprule
        &&\multicolumn{4}{c}{\texttt{ZAM\_Zip}} \\
        \midrule
        Egoistic & - &  7.56 & 416.65 & 92.55 & 9.31\\
        \midrule
        \multirow{3}{*}{Collective} 
            & low & \textbf{7.21} & 406.80 &  94.16 & 9.12 \\
            & mod. & 7.33 & 414.92 & 93.75 & 9.44 \\
            & high  & 7.80 & 427.57 & 91.60 & 9.52\\
            \midrule
        \multirow{3}{*}{Altruistic} 
            & low & 7.39 & \textbf{379.45} & \textbf{99.69} & 7.81 \\
            & mod. & 7.32 &  409.18 & 94.09 & 9.40 \\
            & high & 8.53 & 440.58 & 90.98 & \textbf{9.59} \\
            \midrule
            & &\multicolumn{4}{c}{\texttt{ZAM\_TJunction}} \\
            \midrule
        Egoistic & - & 10.09 & \textbf{1044.13} & \textbf{52.33} & 4.10 \\
        \midrule
        \multirow{3}{*}{Collective} 
            & low & 10.10 & 1050.49 & 51.71 &  \textbf{4.24} \\
            & mod. & 10.20 & 1052.87 & 51.81 & 4.21 \\
            & high &  10.25 & 1054.25 & 52.03 & 4.10 \\
            \midrule
        \multirow{3}{*}{Altruistic} 
            & low & \textbf{10.08} & 1056.05 & 51.20 & 3.96 \\
            & mod. & 10.28 & 1094.95 & 47.76 & 3.84 \\
            & high & 10.41& 1116.01 & 46.39 & 3.61 \\
                        \midrule
            & &\multicolumn{4}{c}{\texttt{USA\_US101}} \\
            \midrule
        Egoistic & - & 9.57 & 650.06 &   108.95 & 5.60 \\
        \midrule
        \multirow{3}{*}{Collective} 
            & low  & 9.55 & 646.54 & 108.23 & 5.57\\
            & mod. & 9.63 & 650.10 & 108.90 & 5.62 \\
            & high & 9.71 & 653.67 & 109.52 & \textbf{5.67}\\
            \midrule
        \multirow{3}{*}{Altruistic} 
            & low &  \textbf{9.16} & \textbf{615.48} & 101.43 & 5.20\\
            & mod. &9.48 & 642.90 & 107.78 & 5.46 \\
            & high & 9.75 & 654.42 & \textbf{110.01} & 5.59\\
        \toprule
    \end{tabularx}
\end{table*}
\section*{Discussion}\label{sec:discussion}
We investigated whether an automated vehicle (AV) motion planning strategy that maximizes travel efficiency and minimizes its own collision risk is adequate in terms of driving ethics in road traffic.
We argued that such an approach falls short in fairly distributing and reducing overall traffic risk, as an AV should also consider the risk from the perspectives of other road users to facilitate ethicality in road traffic.\\
We find that considering the risk perspectives of other road users in a driving scenario has a significant impact on the overall risk in the driving scene and on the AV's driving behavior. Furthermore, the risk as perceived by other road users must generally be considered asymmetric, i.e., different from the risk the AV perceives. In fact, for the AV to accept a small increase in risk, between \qty{7.5}{\%} to \qty{8.7}{\%}, the risk of each other road user can be significantly reduced by \qty{-8.4}{\%} to \qty{-22.3}{\%}, when taking a collective risk perspective. Furthermore, if the AV assesses that other road users perceive its motion as predictable, the AV's and other road users' risks are reduced, improving the AV's travel efficiency by allowing it to act more assertively. Conversely, the AV drives conservatively when it assesses that its motion plan is uncertain to other road users. We consider this a desirable, nuanced behavior by clearly taking other road users' beliefs about the evolution of the driving scene into account. This behavior can be interpreted as a form of \textit{self-reflection} of the AV, which is a natural prerequisite for socially acceptable AV behavior. Such risk allocation for the collective risk perspective is similar to human driving behaviors~\cite{krugel2024risk}, where humans put themselves at at risk for the benefit of others.\\
We conclude that to facilitate ethicality in road traffic, the individual risk-perspectives of each road user in a driving scene must be considered. The resulting benefits in AV behavior should be, besides lowering the total risk, another motivation to consider perspective-based risk in motion planning. Lastly, we believe that for an AV to comply with ethical regulations~\cite{ethicscommission2017, ethicsEU}, perspective-based risk minimization is necessary. 
 
\section*{Methods}\label{sec:methodology} 
To address our objective of balancing maximizing travel efficiency with minimizing risk costs for the three perspective-aware risk costs in (\ref{eq:risk_costs}), expressions for the risk from the ego's perspective $R_{n|k}^{e \gets o}$ and objects' perspectives $R_{n|k}^{o \gets e}$ must be derived. In addition, a motion planning strategy accommodating these risk costs is needed. \\
The following notation will be used throughout this paper. Let any position in $\mathbb{R}^2$ be denoted by $\mathbf{x} := (x, y)$ and any orientation be denoted by $\theta \in \mathbb{R}$. Then, the  configuration space $\mathcal{C} := \mathbb{R}^3$ gives the set of all possible actor (either ego or object) configurations. Note that we define $\theta$ on $\mathbb{R}$, as for later calculations it is beneficial to represent angles as multiples of $2\pi$. We denote a configuration $\mathbf{y} = (\mathbf{x},\theta)$ at discrete-time instant $k$ by $\mathbf{y}_{k}$ and a predicted configuration at discrete-time instant $n \geq k$ given information available at time $k$ by $\mathbf{y}_{n|k}$. The same notation is applied for states, inputs, and constraints. All variables associated with the ego and the objects will be identified, respectively, with $e$ and $o$ subscripts, where $o$ also indexes the objects in the scene, i.e., $o \in \mathbb{Z}_1^{N_o}$. Here, $\mathbb{Z}_a^b$ denotes of set of integers $\{a, a + 1, ..., b \}$, with $ a < b$.

\subsection*{Risk of Collision: Ego Perspective}\label{sec:ego_risk}
\begin{figure}[t!]
\begin{center}
\includegraphics[width=8.4cm]{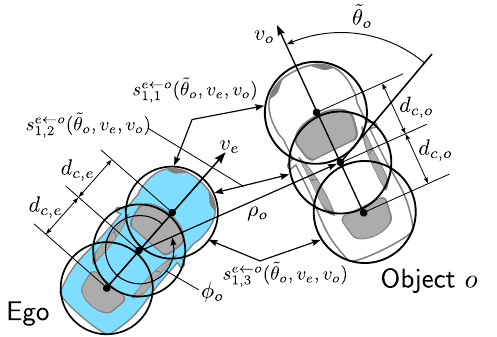}  % The printed column width is 8.4 cm.
\caption{An ego vehicle and object $o$ covering of three circles each. As an example, we depict three severity functions $s_{j,l}$ for a collision between the front ego circle $j = 1$ with each object circle $l$. We omit time indexing for clarity.}
\label{fig:circles}
\end{center}
\end{figure}

The authors of~\cite{Tolksdorf_risk_2025} define the risk for the ego's perspective as the expected severity of collision. This approach is based on a method for collision probability estimation of~\cite{Tolksdorf_POC_2024}, because when the severity term is constant, one is left with estimating the probability of collision. Here, both approaches, i.e, collision probability and risk estimation, utilize a multi-circular shape approximation of each vehicle, which facilitates the derivation of the collision conditions. An illustration for a multi-circular covering is provided in Figure~\ref{fig:circles}.
The risk is estimated given the assumption that only the kinematic variables of the object, i.e., $\mathbf{z}_o= (x_o, y_o, \theta_o, v_o)$, are uncertain to the ego vehicle, and all random variables are mutually independent. Herein, the velocity $v_o$ of object $o$ is directed along its heading $\theta_o$, which is measured with respect to the global $x$-axis. Note that the ego's kinematic variables $\mathbf{z}_e= (x_o, y_e, \theta_e, v_e)$ are defined in the same global coordinate frame. 
To derive the risk, first, the global Cartesian position is transformed into a polar coordinate frame of coordinates relative to the ego vehicle, i.e., 
\begin{equation}\label{eq:polar_transform}
\begin{split}
   & CT: \mathbb{R}^2 \rightarrow \mathbb{R} \times [0, 2\pi),\\
   & (x_e - x_o, y_e - y_o) = (\rho_o \cos{\phi_o}, \rho_o \sin{\phi_o}).
\end{split}
\end{equation} 
Then, for each object $o$, one can derive an upper-bound on the (relative) radial distance $\overline{\rho}_o$  for which collisions can occur~\cite{Tolksdorf_POC_2024}. That distance is determined by the number of circles $N_{c,e}$ and $N_{c,o}$ used for the approximation of the footprint of the ego vehicle and object $o$, respectively, and their placing. For each vehicle, the placing is assumed to be equidistantly along the longitudinal axis of each vehicle, where the circles are placed a distance $d_e$ for the ego, and $d_o$, for an object, apart, see Figure~\ref{fig:circles}. Hence, the radial upper-bound for a collision between the ego vehicle and object $o$ is given by 
\begin{equation}\label{eq:radial_bound}
    \overline{\rho}_o = r_e + r_o + \frac{d_{o}}{2}\big(N_{c,o} - 1\big) + \frac{d_{e}}{2}\big(N_{c,e} -1\big).
\end{equation}
Herewith, for all angles $\phi_o \in [0, 2\pi)$, collisions are possible for $\rho_o \in [0, \overline{\rho}_o]$. We denote $\mathbf{y}_{p,o}$ a polar representation of the object configuration $\mathbf{y}_o$, for which it holds that $\mathbf{y}_{p,o} := (\rho_o, \phi_o, \tilde{\theta}_o) := (\sqrt{(x_e-x_o)^2 + (y_e-y_o)^2}, \text{atan2}(y_e-y_o, x_e-x_o), \theta_e-\theta_o)$, where `$\text{atan2}$' designates the two-argument arctangent function. 
\\
Given the collision conditions in the position coordinates $(\phi_o, \rho_o)$, one is left to consider the relative heading angle $\tilde{\theta}_o$. In~\cite{Tolksdorf_POC_2024}, the collision conditions on the relative heading angle $\tilde{\theta}_o$ are given by a collection of $N_{c,e}N_{c,o}$ intervals $\mathbb{I}^{\tilde{\theta}_o}_{j,l}(\phi_o, \rho_o) = [\underline{\tilde{\theta}}_{j,l,o}(\phi_o, \rho_o), \overline{\tilde{\theta}}_{j,l,o}(\phi_o, \rho_o)]$, for a position $(\phi_o, \rho_o)$, such that if $\tilde{\theta}_o \in \mathbb{I}^{\tilde{\theta}_o}_{j,l}(\phi_o, \rho_o)$ a collision is occurring. Here, the indices $(j,l)$ denote an intersection between the $j$-th ego circle and the $l$-th circle of object $o$.
Hence, the set of all polar collision configurations of object $o$ is given by
\begin{equation}\label{eq:Acir}
\begin{split}
    \mathcal{A}_{p,o} = \{(\phi_o, \rho_o, &\tilde{\theta}_o) \in \mathcal{C}_p  \mid \phi_o \in [0, 2\pi), \rho_o \in [0, \overline{\rho}_o],\\
    &  \tilde{\theta}_o \in \cup_{(j,l) \in \mathbb{Z}_1^{N_{c,e}}\times \mathbb{Z}_1^{N_{c,e}}}\;\mathbb{I}_{j,l}^{\tilde{\theta}_o}(\phi_o, \rho_o) \},
\end{split}
\end{equation}
where $\mathcal{C}_p$ is the polar transformation of the configuration space, i.e., $\mathcal{C}_p := CT^{-1}(\mathbb{R}^2) \times \mathbb{R}$. Similarly, one can apply the polar coordinate transformation to the parameterization of the uncertainty so that $\mathbf{w}^{e\gets o} = (\mathbf{w}^{e\gets o}_{\phi, \rho}, \mathbf{w}^{e\gets o}_{\tilde{\theta}}, \mathbf{w}^{e\gets o}_v)$.  With (\ref{eq:Acir}), one can derive the probability of collision between the ego and object $o$, by integrating probability density functions in the polar configuration, i.e., $p_{\phi, \rho}(\phi, \rho; \mathbf{w}^{e\gets o}_{\phi, \rho})$ for the relative polar position, and $p_{\tilde{\theta}}(\tilde{\theta}; \mathbf{w}^{e\gets o}_{\tilde{\theta}})$ for the relative heading angle, over $\mathcal{A}_{p,o}$.\\
Introducing the severity term,~\cite{Tolksdorf_risk_2025} proposes to derive the severity on a circle-to-circle bases, allowing to take the different severities for different collision constellations, e.g., considering the difference in severity between a head-on or rear-end collision, into account. For example, a rear-end collision can be represented by a specific severity function estimating the severity of a front ego circle colliding with the rear circle of object $o$. Given $N_{c,e}N_{c,o}$ individual severity functions $s_{j,l}^{e\gets o}$, an algorithm is needed that combines all the individual severity functions, related to the combination of a specific ego-to-object circle pair, into a total severity value, as only one risk value is desired. Note that the individual circle-to-circle severity function $s^{e\gets o}_{j,l}$ depends only on the relative heading angle $\tilde{\theta}_o$ and the velocities $v_e, v_o$, as the position is already accounted for by the specific circle-to-circle pair, see Figure~\ref{fig:circles}.\\
To derive the expected collision severity, i.e., risk, the severity function along with the probability density functions in $\phi_o, \rho_o, \tilde{\theta}_o, v_o$, must be integrated over the set of collision configurations, given by (\ref{eq:Acir}). Because the velocities are not needed for collision determination, the integral in the velocity of object $o$ is solved first by integrating over the possible set of object collision velocities at time $n|k$, giving the conditional expected circle-to-circle severity $ \mathbb{E}[s^{e\gets o}_{j,l}(\tilde{\theta}, v_{e}, v_o)\mid v_{o, n|k}]$. Nonetheless, the heading angle is needed for \textit{both}, collision determination (see (\ref{eq:Acir})) and severity estimation, where the individual circle-to-circle severities must be combined over all $\mathbb{I}_{j,l}^{\tilde{\theta}_o}(\phi_o, \rho_o)$ intervals to retrieve a total severity value. Therefore, Algorithm 1 in~\cite{Tolksdorf_risk_2025} decomposes the union of all $\mathbb{I}_{j,l}^{\tilde{\theta}_o}(\phi_o, \rho_o)$ intervals into $N_m$ disjoint intervals $\mathbb{I}_{m}^{\tilde{\theta}_o}(\phi_o, \rho_o)$, and averages $\mathbb{E}[s^{e\gets o}_{j,l}(\tilde{\theta}, v_{e}, v_o)\mid v_{o, n|k}]$ whenever a disjoint interval $\mathbb{I}_{m}^{\tilde{\theta}_o}(\phi_o, \rho_o)$ contains a collision among multiple pairs of circles, which then gives the total conditional expected severity $\mathbb{E}[s^{e\gets o}_m(\tilde{\theta}, v_{e}, v_o) \mid v_{o, n|k}]$. In that sense, the total severity is a weighted sum of individual severity functions $s_m$. With the total conditional expected severity in the velocity over the intervals $\mathbb{I}^{\tilde{\theta}_o}_{m}(\phi, \rho)$, and the coordinate transformation (\ref{eq:polar_transform}), the risk of a collision between the ego vehicle and object $o$ at time $n|k$ from the ego's perspective is given by
\begin{equation}\label{eq:ego_risk_full}
    \begin{split}
        &R^{e \gets o}(\mathbf{y}_{e, n|k}, v_{e, n|k}; \mathbf{w}_{n|k}^{e\gets o})=\\
        &\int_0^{2\pi}\!\int_0^{\overline{\rho}_o}\!p_{\phi, \rho}(\phi, \rho; \mathbf{w}^{e\gets o}_{\phi, \rho, n|k})\Bigg[ \sum_{m=1}^{N_{m}} \!\int_{\mathbb{I}^{\tilde{\theta}_o}_{m}(\phi, \rho)}\\
        & p_{\tilde{\theta}}(\tilde{\theta}; \mathbf{w}_{\tilde{\theta},n|k}^{e\gets o})\mathbb{E}[s^{e\gets o}_m(\tilde{\theta}, v_{e, n|k}, v_o) \mid v_{o, n|k}] \text{d} \tilde{\theta} \Bigg] \text{d}\rho \text{d}\phi,
    \end{split}
\end{equation}
where $\mathbf{w}_{n|k}^{e\gets o} = (\mathbf{w}^{e\gets o}_{\phi, \rho, n|k}, \mathbf{w}^{e\gets o}_{\tilde{\theta}, n|k}, \mathbf{w}^{e\gets o}_{v, n|k})$. For notational clarity we denote the risk from the ego perspective, induced by object $o$ at time $n|k$, as follows from now on:
\begin{equation}\label{eq:ego_risk}
    R^{e\gets o}_{n|k} :=  R^{e \gets o}(\mathbf{y}_{e, n|k}, v_{e,n|k}; \mathbf{w}^{e \gets o}_{n|k}).
\end{equation}

\subsection*{Risk of Collision: Object Perspective}\label{sec:obj_perspective_risk}
The ego's risk $R_{n|k}^{e \gets o}$ in (\ref{eq:ego_risk}) depends on the ego's configuration $\mathbf{y}_{e, n|k}$, the ego's velocity $v_{e, n|k}$, and the ego's uncertainty parameterization $\mathbf{w}_{n|k}^{e\gets o}$ about the object's configuration and velocity. Switching perspectives to derive $R_{n|k}^{o \gets e}$, i.e., the risk from the perspective of object $o$, by replacing the arguments $\mathbf{y}_{e,n|k}, v_{e,n|k}$ and $\mathbf{w}_{n|k}^{e\gets o}$ with $\mathbf{y}_{o,n|k}, v_{o,n|k}$ and $\mathbf{w}_{n|k}^{o\gets e}$, neglects the fact that $\mathbf{y}_{o,n|k}$ and $v_{o,n|k}$ are, by assumption (see Assumption 1 in Supplementary Note 1), uncertain to the ego.
A simple and yet effective way to circumvent this issue is to model the object risk $R_{n|k}^{o \gets e}$ to be dependent on the uncertainties $\mathbf{w}_{n|k}^{o\gets e}$ that the object $o$ has about the ego's configuration and velocity only, and to use the object mean values for $\mathbf{y}_{o,n|k}$ and $v_{o,n|k}$, i.e., $\mu_{\mathbf{y}_o,n|k} = (\mu_{x_o,n|k}, \mu_{y_o,n|k}, \mu_{\theta_o,n|k})$ and $\mu_{v_o,n|k}$ as arguments to $R_{n|k}^{o \gets e}$. Essentially, this reflects that the object does not have uncertainty about its own variables, and the underlying assumption is that the mean object trajectories available to the ego are unbiased. Note that the uncertainty that the object has about the kinematic variables of the ego is still taken into account in $R^{o\leftarrow e}_{n|k}$. Therefore, to construct the risk from the object's perspective, the risk derivation for the ego's perspective is adapted by changing $e \gets o$ to $o \gets e$ and the ego configuration $\mathbf{y}_{e,n|k}$ and velocity $v_{e,n|k}$ are changed to the object means $\mu_{\mathbf{y}_o,n|k}$ and $\mu_{v_o,n|k}$, respectively. As such, with (\ref{eq:ego_risk}), the risk function for object $o$ can be defined as follows:
\begin{equation}\label{eq:object_risk}
    R_{n|k} ^{o \gets e} := R^{e \gets o}(\mu_{\mathbf{y}_{o,n|k}}, \mu_{v_o,n|k}; \mathbf{w}^{o \gets e}_{n|k}).
\end{equation}
\subsection*{Risk-based Motion Planning}\label{sec:opt_motion_planning}
In this section, we will formulate an optimization problem balancing path-following with risk cost minimization, while taking the different risk perspectives into account.
Let the motion dynamics of the ego  be modeled by a discrete-time non-linear system (e.g., obtained by time discretization of a continuous-time vehicle dynamics model) of the form:
\begin{equation}\label{eq:sys}
\begin{split}
\mathbf{q}_{k+1} &= f(\mathbf{q}_k, \mathbf{u}_{k}), \\
(\mathbf{y}^T_{e, k}, v_{e,k}) & = h(\mathbf{q}_{k}),
\end{split}
\end{equation}
where $\mathbf{q}_k \in \mathbb{R}^{n_s}$ represents the ego's state vector and $\mathbf{u}_{k} \in \mathbb{R}^{n_{u}}$ is the input vector at time $k$. Furthermore, $\mathbf{y}_{e, k}$ and $v_{e,k}$ are the ego's configuration and velocity, directed along its heading axis, at time $k$, respectively. The model equations are represented by $f$, and $h$ is the output function. State and input constraints are given by bounding $\mathbf{q}_k$ and $\mathbf{u}_{k}$ to specific sets, i.e., $\mathbf{q}_k \in \mathcal{Q}$ and $\mathbf{u}_{k} \in \mathcal{U}$ for all $k$, respectively. Suppose that the ego vehicle is following time-independent references, where we denote the set of reference configurations as a path $\mathcal{P}$ and the reference velocity as $v_{ref}$. These are, e.g., the center of the right-most lane and a local speed limit, eventually leading to the ego's goal destination. The path $\mathcal{P}$ is a regular curve parameterized by $\lambda$ over $[\lambda_0, \lambda_g]$, see~\cite{Tolksdorf_POC_2024}. Path-following treats the progress along the path as an additional degree of freedom to the controller and, therefore, the progress along the path, i.e., the evolution of $\lambda_{n|k} \in [\lambda_0, \lambda_g]$, depends on the control input $\mathbf{u}_{n|k}$. Note that $\lambda_{n|k}$ reflects the configuration of the ego vehicle along the path. As the ego is performing recursive planning, i.e., at each time-instance a new trajectory is computed and the first control action is applied to the vehicle, the initial reference point on the path at each planning cycle is determined by finding the closest point on the path $\mathcal{P}$ to the ego configuration, i.e.,
\begin{equation} \label{eq:init_input}
\lambda_{k|k} = \argmin_{\lambda \in [\lambda_0, \lambda_g]} \| \mathbf{y}_{e, k|k} - \mathbf{y}_P(\lambda) \|, 
\end{equation}
where $\mathbf{y}_P(\lambda) \in \mathcal{P}$. Within the planning horizon, we use the ego's velocity directed along the ego's heading to approximate the progress along the path with 
\begin{equation}\label{eq:input}
\lambda_{n+1|k} = \lambda_{n|k} + v_{e,n|k}\cos{(\theta_{e,n|k} - \theta_{p,n|k})}\Delta T, 
\end{equation}
for $n \in \mathbb{Z}_{k+1}^{k + N_P -1}$, where $\theta_p$ is the reference heading angle, i.e., the angle of the tangent to the path at $\lambda_{k|n}$ (see Figure 2 in~\cite{tolksdorf2023risk}) and $\Delta T$ represents the sample interval associated to the discrete-time vehicle dynamics model in~(\ref{eq:sys})\footnote{Note that the accuracy of the approximation (\ref{eq:input}) diminishes as the path's curvature increases.}. Clearly, the definition of the references does not account for the presence of objects and their motion. As a consequence, the ego vehicle may potentially have to deviate away from its reference to minimize risk and, therewith, shape its motion around objects. Thus, the ego aims to balance path-following error minimization with risk minimization over a prediction horizon $\mathbb{Z}_k^{k+N_P}$, where the reference error is given by 
\begin{equation}\label{eq:ref_error}
    \begin{split}
    \mathbf{e}(\lambda_{n|k}) & =
    \begin{pmatrix}
    \mathbf{y}_{e,n|k}^T - \mathbf{y}_P^T(\lambda_{n|k})\\
    v_{e,n|k} - v_{ref}
    \end{pmatrix}.
\end{split}
\end{equation}
To minimize the path-following error, we define a cost function $J_P$ which quadratically penalizes deviations of the reference path and reference velocity as 
\begin{equation}\label{eq:ref_costs}
    J_P(\lambda_{n|k})= \mathbf{e}(\lambda_{n|k})^T W\mathbf{e}(\lambda_{n|k})
\end{equation}
with a positive definite weighting matrix $W \in \mathbb{R}^ {4\times4}$.
Herewith, the ego vehicle is tasked with finding the input sequence $\mathbf{U}_k := [\mathbf{u}_{k|k}, \mathbf{u}_{k+1|k}, ..., \mathbf{u}_{k+N_P|k}]$ for a prediction horizon of length $N_P \in \mathbb{N}_{>0}$ by recursively solving the following optimization problem:
\begin{subequations}\label{eq:general_OP}
\begin{equation}
   V^{SMPC}(\mathbf{q}_{k}, \lambda_k) := \min_{\mathbf{U}_k} [J_{r, k} +\sum_{n = k}^{k + N_P} J_P(\lambda_{n|k})],\label{eq:smpc_a}
\end{equation}
subject to:
\begin{align}
    & \quad \mathbf{q}_{k|k} = \mathbf{q}_k, \;\;\;\; \lambda_{k|k} = \lambda_k,\label{eq:smpc_b} \\
& \forall n \in \mathbb{Z}_k^{k+N_P}: \mathbf{q}_{n+1|k} = f(\mathbf{q}_{e, n|k}, \mathbf{u}_{n|k}),  \label{eq:smpc_c}\\
&\quad (\mathbf{y}_{e,n|k}^T, v_{e, n|k})  = h(\mathbf{q}_{n|k})\label{eq:smpc_d},\\
&\quad \lambda_{n+1|k}\!=\!\lambda_{n|k} + v_{e,n|k}\cos{(\theta_{e,n|k} - \theta_{p,n|k})}\Delta T,\label{eq:smpc_e}\\
&\quad \mathbf{e}(\lambda_{n|k})= (\mathbf{y}_{e,n|k}^T - \mathbf{y}_P^T(\lambda_{n|k}), v_{e,n|k} - v_{ref})^T,\label{eq:smpc_f} \\
&\quad (\mathbf{q}_{n|k}, \mathbf{u}_{n|k}, \lambda_{n|k}) \in \mathcal{Q} \times \mathcal{U} \times [\lambda_0, \lambda_g],\label{eq:smpc_g}\\
&\quad J_{r,k}\text{ being one of the risk costs in (\ref{eq:risk_costs}).\label{eq:smpc_h}}
\end{align}
\end{subequations}
Here, the initial conditions are set in (\ref{eq:smpc_b}). The system model is provided in (\ref{eq:smpc_c}) - (\ref{eq:smpc_e}), from which the error is calculated in (\ref{eq:smpc_f}) to support path-following by minimization of this error with (\ref{eq:ref_costs}). State and input constraints are given in (\ref{eq:smpc_g}). Lastly, (\ref{eq:smpc_h}) implements the user-selected risk costs from (\ref{eq:risk_costs}). 
\subsection*{Evaluation Metrics}\label{sec:sim_metrics}
In addition to recording the risk costs, we also record the ego's driving by the following measures: 
\begin{enumerate}[leftmargin=0.5cm]
    \item \textit{Distance to objects:} The Euclidean distance between the ego vehicle's geometric center and the center of all object vehicles in the scene at time $k$.
    \item \textit{Reference error:} We record the accumulated reference error as $\sum_{k=0}^{N_T}||\mathbf{e(\lambda_{k|k})}||$, where $N_T$ is the number of time steps in a scenario, see (\ref{eq:ref_error}). Furthermore, we record the maximum reference error given by $\max_k||\mathbf{e(\lambda_{k|k})}||$.
    \item \textit{Traveled distance:} We record the traveled ego vehicle distance as $\sum_{k=1}^{N_T}|| \mathbf{x}_{e,k} - \mathbf{x}_{e,k-1}||$.
\end{enumerate}
As we are also interested in understanding the risk perspective's impact on the scenario cluster and overall traffic risk, we define the following evaluation metrics:
\begin{enumerate}[leftmargin=0.5cm]
    \item \textit{Average accumulated risk cost:} We define the accumulated risk cost at time $k$ for a given scenario $s$ in scenario cluster $c$ as $J^{s,c}_{acc,k} = \sum_{i=0}^kJ_{r,i}$, where $r$ denotes any of the three risk costs from (\ref{eq:risk_costs}) at time $i$ for $0 \leq i \leq k$. Then, the average accumulated risk at time $k$ for scenario cluster $c$ is calculated as $\frac{1}{N^c_s}\sum_{s=1}^{N_{s}^c}J^{s,c}_{acc,k}$, where $N^c_s$ is the number of scenarios in scenario cluster $c$.
    \item \textit{Total weighted average risk cost:} We define the average risk cost per scenario $s$ in scenario cluster $c$ as $J^{s,c}_{r,avg} = \frac{1}{N_T}\sum_{k=0}^{N_T} J_{r,k}$. Then, the total weighted average risk cost is given by $\frac{1}{N_{s,tot}}\sum_{c\in clusters }(N^c_s\sum_{s=1}^{N^c_s}J^{s,c}_{r,avg})$, where $N_{s,tot}$ is the total number of scenarios over all scenario clusters, i.e., $683$ in this paper.
    \item \textit{Total weighted average maximum risk cost:} We define the maximum weighted risk cost per scenario $s$ in scenario cluster $c$ as $J^{s,c}_{r,max} = \max_{k} J_{r,k}$. The total weighted average maximum risk cost is given by $\frac{1}{N_{s,tot}}\sum_{c\in clusters}(N^c_s\sum_{s=1}^{N^c_s}J^{s,c}_{r,max})$.
\end{enumerate}
\section*{Supplementary Note 1: Problem Statement}
Consider an arbitrary traffic scene with an automated (ego) vehicle and $N_o$ other actors (e.g., vehicles or cyclists), referred to as the objects. We characterize each actor, i.e., ego and objects, by a configuration composed of a position and orientation. Let any position in $\mathbb{R}^2$ be denoted by $\mathbf{x} := (x, y)$ and any orientation be denoted by $\theta \in \mathbb{R}$. Then, the  configuration space $\mathcal{C} := \mathbb{R}^3$ gives the set of all possible actor configurations. Note that we define $\theta$ on $\mathbb{R}$, as for later calculations it is beneficial to represent angles as multiples of $2\pi$. We denote a configuration $\mathbf{y} = (\mathbf{x},\theta)$ at discrete-time instant $k$ by $\mathbf{y}_{k}$ and a predicted configuration at discrete-time instant $n \geq k$ given information available at time $k$ by $\mathbf{y}_{n|k}$. The same notation is applied for states, inputs, and constraints. All variables associated with the ego and the objects will be identified, respectively, with $e$ and $o$ subscripts, where $o$ also indexes the objects in the scene, i.e., $o \in \mathbb{Z}_1^{N_o}$. Here, $\mathbb{Z}_a^b$ denotes of set of integers $\{a, a + 1, ..., b \}$, with $ a < b$.

\subsection*{Perspective-based Risk}\label{sec:perspective_based_risk}
We adopt the risk definition from~\cite{Tolksdorf_risk_2025}, where risk is defined as the expected severity of a collision. Herein, a risk measure is constituted of two aspects and considers the ego's perspective only: the first aspect considers the ego's perception and motion prediction uncertainty associated with the kinematic variables (i.e., configuration and velocity) of an actor, and the other aspect estimates the severity of a potential collision between the ego and the said actor (see also Figure 1(a) of the main manuscript for an illustrative example of the risk defining aspects).\\
In ~\cite{Tolksdorf_risk_2025}, Eq. (3), the risk is estimated between the ego vehicle and an object vehicle, by using the ego's and object's kinematic variables given by their configurations and velocities (directed along their heading axis), i.e., $\mathbf{z}_e = (x_e, y_e, \theta_e, v_e)$ and $\mathbf{z}_o= (x_o, y_o, \theta_o, v_o)$, respectively. Hereto, we assume that the ego vehicle is equipped with sensors, perception and prediction algorithms to estimate the current and future kinematic variables of all sensed objects. Due to the inherent uncertainty in measurement and estimation, we indeed consider $\mathbf{z}_{o}$ to be uncertain. Considering such uncertainty, we let $R^{e \gets o}_k \in \mathbb{R}_{\geq 0}$ capture the risk of collision of the ego vehicle with object $o$ from the ego vehicle's perspective. Therefore, $R^{e \gets o}_k$ encompasses the ego vehicle's uncertainty about the object's kinematic variables $\mathbf{z}_o$, represented by the parameters $\mathbf{w}^{e \gets o}_k$ of the uncertainty model, and the ego's potential severity of collision $s^{e \gets o}\in \mathbb{R}_{\geq 0}$ with the object at discrete-time instance $k$. As $R^{e \gets o}_k$ only reflects the risk the object $o$ poses to the ego vehicle, we denote $R^{o \gets e}_k \in \mathbb{R}_{\geq 0}$ as the risk the ego poses to the object vehicle $o$. To estimate $R^{o \gets e}_k$, the ego vehicle is tasked to estimate the uncertainty the object faces about the ego's kinematic variables $\mathbf{z}_e$, i.e., taking the perception and prediction perspective of the object. In addition, the ego vehicle is tasked to estimate the severity function from the object's perspective, i.e., $s^{o \gets e}\in \mathbb{R}_{\geq 0}$. To estimate the risks $R^{e \gets o}_k$ and $R^{o \gets e}_k$ we adopt the following assumptions, which express which information is available to the ego vehicle's motion planning algorithm.

\begin{assumption}(Ego perspective)\label{ass:object_info}
    For each object $o \in \mathbb{Z}_1^{N_o}$, the ego vehicle estimates the kinematic variables $\mathbf{z}_{o,n|k}$ at discrete-time instance $n \geq k$, given information available at time $k$, together with their associated probability density function $p_{\mathbf{z}_{o}}$, parametrized by the time-varying vector of uncertainty model parameters $\mathbf{w}_{n|k}^{e\gets o}$. The parametrization $\mathbf{w}_{n|k}^{e\gets o}$ is estimated for all instances $n$ within a prediction horizon $\mathbb{Z}_k^{k+N_P}$ with a prediction model. All components in $\mathbf{z}_o$ are mutually independent for all $n$. 
\end{assumption}

\begin{assumption}(Estimated object perspective)\label{ass:ego_info}
    From the perspective of each object $o \in \mathbb{Z}_1^{N_o}$, the ego vehicle estimates the uncertainty each object faces about the ego's kinematic variables $\mathbf{z}_{e,n|k}$ at discrete-time instance $n\geq k$, given information available at time $k$, by their associated probability density function $p_{\mathbf{z}_e}$, parametrized by the time-varying vector of uncertainty model parameters $\mathbf{w}_{n|k}^{o\gets e}$. The parametrization $\mathbf{w}_{n|k}^{o\gets e}$ is estimated for all instances $n$ within a prediction horizon $\mathbb{Z}_k^{k+N_P}$ with a prediction model. All components in $\mathbf{z}_e$ are mutually independent for all $n$. 
\end{assumption}

\begin{assumption}(Risk induced by interaction between objects)\label{ass:no_cross_obj_risk}
    The motion of the ego vehicle does not affect the risk between objects, and, thus, the risk between object vehicles is not estimated. 
\end{assumption}
\subsection*{Behavior Generation}
Suppose the ego vehicle is provided with reference signals, guiding it towards a desired destination and it is planning its trajectory by deriving a sequence of control actions (e.g., steering and acceleration) $\mathbf{U}_k := [\mathbf{u}_{k|k}, \mathbf{u}_{k+1|k}, ..., \mathbf{u}_{k+N_P|k}]$. Hence, at each time step within the ego's planning horizon $\mathbb{Z}_k^{k+N_P}$, it estimates the error between its configuration and velocity and the associated reference signals. Using these error signals, the cost $J_{P, n|k} \in \mathbb{R}_{\geq 0}$ is calculated, such that $J_{P, n|k} = 0$ if the ego follows the reference exactly at planned time instance $n$. Therefore, to generate its trajectory, the ego strives to minimize $J_{P, n|k}$ along its planning horizon.\\
Besides this, another objective is to minimize risk. In order to generate behavior based not only on the ego's risk $R^{e\gets o}_{n|k}$, but also on the risk of objects $R^{o\gets e}_{n|k}$, the object's risk perspective must be dependent on the ego's input $\mathbf{u}_{n|k}$. Otherwise, the object risks $R^{o\gets e}_{n|k}$ would be independent of the ego's planned control actions and, therefore, could not be minimized by optimizing for these control actions. To take this into account, not only in the severity term, but also in the uncertainty term, we adopt the following assumption.
\begin{assumption}\label{ass:controllable_object_perspective}(Availability of object-perspective uncertainty model)
    The ego vehicle estimates the uncertainty $\mathbf{w}_{n|k}^{o\gets e}$ each object faces about the ego's kinematic variables $\mathbf{z}_{e,n|k}$ by a prediction model $\Pi$ mapping the ego's predicted kinematic variables $\mathbf{z}_{e,n|k}$ on the object's uncertainty parameterization of the ego's kinematic variables as follows:
    \begin{equation}\label{eq:perspective}
        \Pi : \mathbf{z}_{e,n|k} \mapsto \mathbf{w}^{o\gets e}_{n|k}.
    \end{equation}
\end{assumption}
Assumption \ref{ass:controllable_object_perspective} can be interpreted as the capacity for self-reflection of the ego vehicle in terms of assessing how predicable an action of itself would be to another actor (Note that an illustration of the Assumptions~\ref{ass:object_info} -~\ref{ass:controllable_object_perspective} is depicted in Figure 1(b) of the main manuscript).\\
Given the Assumption~\ref{ass:controllable_object_perspective}, we can distinguish three cases of risk costs to be minimized:
\begin{equation}\label{eq:risk_cases}
\begin{split}
    &J_{r,n|k} = \\
    &\begin{cases}
         \frac{w_R}{N_o}\sum_{o = 1}^{N_o} \gamma(n-k)R_{n|k}^{e \gets o} \triangleq \text{egoistic risk cost}, \\
         \frac{w_R}{2N_o} \sum_{o = 1}^{N_o}\gamma(n-k)(R_{n|k}^{e \gets o} + R_{n|k}^{o \gets e})  \triangleq \\ \quad\quad\quad\quad\quad\quad\quad\quad\quad\quad\quad \text{collective risk cost},\\
         \frac{w_R}{N_o} \sum_{o = 1}^{N_o} \gamma(n-k)R_{n|k}^{o \gets e} \triangleq \text{altruistic risk cost}, 
    \end{cases}
\end{split}
\end{equation}
where $w_R \in \mathbb{R}_{>0}$ is a weighting scalar and $\gamma: \mathbb{Z}_{\geq 0} \rightarrow (0, 1]$ is a function that discounts risk over the prediction horizon ($n-k$ gives the current prediction step). \\
The problem considered in this paper is to develop a methodology that supports finding the input sequence $\mathbf{U}_k := [\mathbf{u}_{k|k}, \mathbf{u}_{k+1|k}, ..., \mathbf{u}_{k+N_P|k}]$ for the ego vehicle, given a prediction horizon of length $N_P \in \mathbb{N}_{>0}$, by recursively solving an optimization problem that balances path-following with risk, i.e., that balances $J_{P, n|k}$ and $J_{R, n|k}$ with the latter representing one of the perspective-based risk costs in (\ref{eq:risk_cases}).
\section*{Supplementary Note 2: Numerical Experiments}
We demonstrate the behavior of the proposed motion planning strategy given different risk-perspectives (see (\ref{eq:risk_cases})), to examine the effects of asymmetric risk on the ego's behavior and overall traffic risk. To this end, we use the CommonRoad \cite{althoff2017commonroad} framework and scenario database, which is widely used to evaluate motion planning strategies, to perform large-scale, simulation-based case studies.
\subsection*{Road User Motion Prediction}\label{sec:prediction}
An essential premise in our work (see Assumptions \ref{ass:object_info} and \ref{ass:ego_info}) is the need for a road user prediction model that evolves uncertainties associated with each road user's configuration and velocity forward in time. That is, a motion prediction model estimating each road user's uncertainty parameterization $\mathbf{w}^{e\gets o}_{n|k}$ and $ \mathbf{w}^{o\gets e}_{n|k}$ for all $n$ in the prediction horizon $\mathbb{Z}_{k}^{k + N_p}$. Hereto, we use the motion prediction model \texttt{WaleNet} from \cite{geisslinger2021watch} which is trained to specifically work with CommonRoad. During our testing, we noticed that \texttt{WaleNet} often gives physically impossible predictions, that are, particularly in the first time instances, overly confident. To account for that, whenever we detected that a prediction made by \texttt{WaleNet} exceeded a given acceleration limit, we used a constant velocity and turn rate (CTRV) model instead. To retrieve uncertainty from the CTRV prediction mode, we  propagated an initial uncertainty $\Sigma_{0} =(\sigma_{x_{0}}, \sigma_{y_{0}})$, which was assumed to grow linearly over the prediction horizon, with an additive term $Q \in \mathbb{R}^{2}$, i.e., $\Sigma_{n|k} = \Sigma_{0} + (n-k)Q$.\\
The resultant \textit{hybrid} motion prediction model provides the means $\mu_x, \mu_y$ and variances $\sigma_x^2, \sigma_y^2$ of the position coordinates for each road user (i.e., also for the ego). As we assume that all random variables are mutually independent, we use a method for uncertainty propagation to estimate $\mu_v, \sigma_v^2$ and $\mu_\theta, \sigma_\theta^2$ (see \cite{kappaupsilon1966notes}, Eq. 2.1 and 2.2) from $\mu_x, \sigma_x^2$ and $\mu_y, \sigma_y^2$. Herewith, we retrieve for the velocity and orientation, respectively,
\begin{equation}
\begin{split}
    &\mu_v = \sqrt{\Delta \mu_x^2 + \Delta \mu_y^2}, \; \sigma_v^2 = \left(\frac{\Delta \mu_x}{\mu_v}\right)^2 + \left(\frac{\Delta \mu_y}{\mu_v}\right)^2, \\
    &\mu_{\theta} = \text{atan2}(\mu_y, \mu_x), \; \sigma_{\theta}^2 = \frac{1}{(\mu_x + \mu_y)^2}(\mu_x^2\sigma_y^2 + \mu_y^2\sigma_x^2),
\end{split}
\end{equation}
where $\Delta\mu_x$ denotes the discrete-time derivative of $\mu_x$. For the purpose of risk estimation from the ego's perspective, we collect the uncertainty parameterization as 
\begin{equation}\label{eq:ego_uncertainty}
    \mathbf{w}^{e\gets o}_{n|k} := \begin{pmatrix}
        (\mu^{o}_{x,n|k}, \sigma^{o}_{x, n|k})\\
        (\mu^{o}_{y, n|k}, \sigma^{o}_{y, n|k})\\
        (\mu^{o}_{\theta, n|k}, \sigma^{o}_{\tilde{\theta}, n|k})\\
        (\mu^{o}_{v,n|k}, \sigma^{o}_{v, n|k})
    \end{pmatrix}.
\end{equation}
Note that the superscripts $o$ denote the respective identifier of the object in the scene. A consequence of the global coordinate prediction is that each object has the same uncertainty prediction about the ego, whereas the ego has individual uncertainty predictions for each object\footnote{We note that current literature regarding prediction models accounting for the objects' perspectives, i.e., a combination of the models prescribed by Assumptions~\ref{ass:ego_info} and~\ref{ass:controllable_object_perspective}, is in its infancy, see, e.g.,~\cite{gil2024predictability}. However, ethicality requirements~\cite{ethicscommission2017, ethicsEU} may make those a necessity, given that risk is indeed asymmetric, and considering it supports a fair distribution and minimization of overall traffic risk. Therefore, we construct such a model out of a global coordinate prediction of all road users as a means to analyze and understand the consequences of asymmetric risk.}.
To cover a wider range of possible object-to-ego predictions, we introduce different uncertainty cases for the objects' perspectives on the ego vehicle. Therefore, for each object's perspective on the uncertainty on the motion of the ego, parametrized by $\mathbf{w}^{o\gets e}_{n|k}$, we introduce a positive scalar weighting $a$, such that\footnote{Note that, in the analysis below, we assign the same weighting $a$ for all objects within a scenario, otherwise the number of different scenarios would  become computationally intractable to simulate as some scenarios contain more than 20 objects (see Section~\ref{sec:scenarios}).} 
\begin{equation}\label{eq:object_uncertainty}
    \Pi (\mathbf{z}_{e,n|k}) = \mathbf{w}^{o\gets e}_{n|k} := \begin{pmatrix}
        (x_{e, n|k}, a\sigma^{e}_{x, n|k})\\
        (y_{e, n|k}, a\sigma^{e}_{y, n|k})\\
        (\theta_{e, n|k}, a\sigma^{e}_{\tilde{\theta}, n|k})\\
        (v_{e,n|k}, a\sigma^{e}_{v, n|k})
    \end{pmatrix},
\end{equation}
where $\mathbf{z}_{e,n|k} = (x_{e, n|k}, y_{e, n|k}, \theta_{e, n|k}, v_{e,n|k})$ is given by eq. (8) in the main paper, and under Assumption \ref{ass:controllable_object_perspective} mapped on $\mathbf{w}^{o\gets e}_{n|k}$.
Hence, the weighting factor allows us to tune the uncertainty of all object's about the ego's future motion. We distinguish $0 < a < 1$ as \textit{low uncertainty}, $a = 1$ as \textit{moderate uncertainty}, and $1 < a$ as \textit{high uncertainty}. Lastly, we bound each $a\sigma$ from below and above, i.e., $\underline{\sigma} \leq a\sigma \leq \overline{\sigma}$, to avoid numerical instabilities\footnote{For computational efficency, our algorithm numerically integrates a bi-variant Gaussian probability density function (PDF) over a pre-sampled grid of points. Small variances cause the non-zero values of Gaussian densities to be closely centered around the mean. Hence, when the mean value is not in close proximity to a pre-sampled point, the PDF of the Gaussian is not evaluated at points where its functional value is non-zero, causing the integral of the PDF to become zero (essentially, the grid, i.e., the numerical precision of the integration, has become too coarse). This will cause the risk to be zero as well, even if a collision occurs.}.  
\subsection*{Perspective-based Risk}
To estimate the perspective-based risk, we use an implementation for Gaussian uncertainties with a kinematic energy collision severity model, see~\cite{Tolksdorf_risk_2025}. We note that the kinematic severity model is independent of the relative heading angle $\tilde{\theta}$ and only depends on the circle-to-circle pair of the object and ego involved in the collision as well as their velocities. Therefore, $s^{e\gets o}_{j,l}(\tilde{\theta}, v_e, v_o)$ becomes $s^{e\gets o}_{j,l}(v_e, v_o)$ and the conditional expected circle-to-circle severity thereof becomes $\mathbb{E}[s^{e\gets o}_{j,l}(v_{e}, v_o)\mid v_{o, n|k}]$, where the total conditional expected severity value then reads $\mathbb{E}[s^{e\gets o}_{m}(v_e, v_o)\mid v_{o, n|k}]$, see eq. (5) in the main paper. For the ego's perspective, the usage is straightforward with (\ref{eq:ego_uncertainty}); however, for the objects' perspectives, a slight modification is necessary under Assumption \ref{ass:controllable_object_perspective}. 
Recall, the ego assumes that its planned trajectory is the mean expected trajectory from the object's perspective, which results in (\ref{eq:object_uncertainty}). However, due to the structure of the kinematic energy model (see Eq. (21) in~\cite{Tolksdorf_risk_2025}), the severity function behaves very differently from the object's perspective compared to the ego perspective, given Assumption \ref{ass:controllable_object_perspective}, thus undermining the accuracy of the comparison of both perspectives. Therefore, instead of having $v_{e,n|k}$ in (\ref{eq:object_uncertainty}) (last row), we use the mean object velocity $\mu^{o}_{v,n|k}$ and estimate the object risk with $ R_{n|k} ^{o \gets e} := R^{e \gets o}(\mu_{\mathbf{y}_{o,n|k}}, v_{e,n|k}; \mathbf{w}^{o \gets e}_{n|k})$. While we still apply the uncertainty levels with $a$ to $\sigma^{e}_{v, n|k}$ from the object's perspective, this causes the change in the perspective in $v$ to be less impactful than for $x, y, \theta$. Hence, given a suitable severity model, we expect the risk to become more asymmetric when the change of perspectives would be applied in its entirety to $v$.

\subsection*{Scenarios and Planner Setup}\label{sec:scenarios} 
\begin{figure}[t!]
\begin{center}
\includegraphics[width=8.4cm]{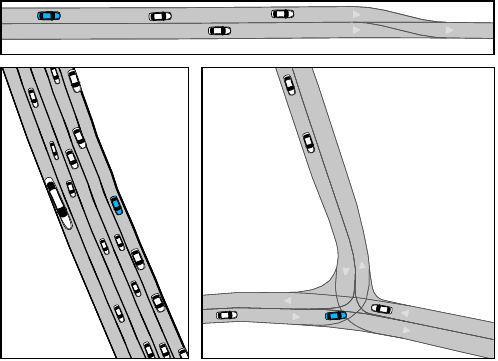}  % The printed column width is 8.4 cm.
\caption{Scenario clusters. Top: \texttt{ZAM\_Zip}, bottom left: \texttt{USA\_US101}, bottom right: \texttt{ZAM\_Tjunction}. The blue depicted vehicle represents the ego.}
\label{fig:scenario_clusters}
\end{center}
\end{figure}

The CommonRoad dataset \cite{althoff2017commonroad} contains a wide range of real-world and hand-crafted scenarios, where we pick three scenario clusters. The specifics of each scenario cluster are provided in Table \ref{tab:scenarios}. Here \texttt{ZAM\_Zip} and \texttt{ZAM\_Tjunction} are hand-crafted scenario clusters, where \texttt{ZAM\_Zip} represents merging scenarios and  \texttt{ZAM\_Tjunction} T-junction scenarios. Lastly, \texttt{USA\_US101} contains recorded real-world scenarios from the NGSIM~\cite{NGSIM} dataset. Exemplary depictions of each scenario cluster are given in Figure \ref{fig:scenario_clusters}.
\begin{table}
    \centering
    \caption{Specifics of CommonRoad scenario clusters.}
    \label{tab:scenarios}
    \begin{tabularx}{\linewidth}{L | C |  P{2.2cm} | C}
        \toprule
        &\texttt{ZAM\_Zip}& \texttt{ZAM\_Tjunction} & \texttt{USA\_US101}  \\
        \toprule
        No. of scenarios &  71 & 547 &65 \\
        Varying road geometry &\xmark &\xmark & \cmark \\
        Varying no. of obj.  &\xmark &\xmark & \cmark\\
        Varying init. cond. of obj.  &\cmark &\cmark & \cmark\\
        Reference velocity [m/s] &15&10&10\\
        \toprule
        %\bottomrule
    \end{tabularx}
\end{table}
Throughout each scenario cluster, we set a constant reference velocity and use reference waypoints leading to a goal destination, which are provided by CommonRoad. To the waypoints, we fit a 3rd-order polynomial to retrieve the reference path, similar to ~\cite{mcnaughton2011motion}. 
To achieve realistic ego vehicle behaviors, the objective function in the optimization problem in (see eq. (13a) in the main paper) must be augmented to also accommodate the road network (e.g., staying in lane) and comfort requirements. Hence, we use an artificial potential field (APF) to reflect the static road network. Here, we use the method from \cite{wolf2008artificial}; however, we only use the road potential $J_{road}$ to represent uncrossable road boundaries and the lane potential $J_{lane}$ to represent crossable lane markings. Using the same method as for the path, we fit a 3rd-order polynomial across a set of points representing a lane marker or road boundary and estimate the nearest point to the lane marker or road boundary with eq.
(9) and (10) of the main paper. Note that each road boundary and lane marker use a distinct parameterization (denoted by $\lambda$ in eq. (9) and (10) of the main paper), which, for clarity of notation, are all collected in a vector $\Lambda$. From the nearest point to lane markers and road boundaries we calculate the APF cost for all roads and lanes as follows: 
\begin{equation}
\begin{split}
    &J_{APF}(\Lambda_{n|k}) = \\
    &W_{APF}\left[\sum_{roads} J_{road}(\Lambda_{n|k}) + \sum_{lanes}J_{lane}(\Lambda_{n|k})\right]^2,
\end{split}
\end{equation} where $W_{APF}$ is a scalar positive weighting. While comfort is not a strict requirement in the problem statement, we still place cost on control effort, to smooth out the ego's behavior. Given a discrete-time unicycle model to represent the ego's motion dynamics (see \cite{tolksdorf2023risk}{, Equation (10)}), with the inputs $\mathbf{u}_{n|k} = (v_{e, n|k}, \Delta\theta_{e, n|k})$ we place cost on the controls with
\begin{equation}
J_{\mathbf{u}}(\mathbf{u}_{n|k}) = \begin{pmatrix}
    \Delta v_{e, n|k}\\
    \Delta\theta_{e, n|k}
\end{pmatrix}^TW_{ctrl}\begin{pmatrix}
    \Delta v_{e, n|k}\\
    \Delta\theta_{e, n|k}
\end{pmatrix},
\end{equation} where $W_{ctrl}\in \mathbb{R}^{2\times2}$ is a positive definite weighting matrix. 
Given $J_{APF}, J_{\mathbf{u}}$ and $\gamma$, we extend the cost function from eq. (13a) of the main paper to
\begin{equation}
    J = \sum_{n = k}^{k+N_p}[J_P(\lambda_{n|k}) + J_{R, n|k} + J_{APF}(\Lambda_{n|k}) + J_{\mathbf{u}}(\mathbf{u}_{n|k})].
\end{equation}
Lastly, we define a discount function $\gamma$, that discounts risk over time, here we choose
\begin{equation}\label{eq:gamma}
    \gamma(n-k) = \frac{1}{N_P}\exp\left[\frac{c_d(n-k)}{N_P}\right],
\end{equation}
where $c_d$ is a positive scalar that adjusts the discount amount. 
\section*{Supplementary Table 1}
The standard deviations of all scenario clusters are reported in Supplementary Table~\ref{tab:results_risk_distribution}.
\refstepcounter{supptable}
\label{tab:results_risk_distribution}
\begin{table}
\centering
    \caption*{Supplementary Table \thesupptable: Standard deviation of accumulated collective risk cost per scenario cluster. Lowest values per scenario cluster and uncertainty setting are highlighted in bold font.}
    
    \begin{tabularx}{\linewidth}{ C | C | C | C}
        \toprule
        Uncertainty Level & Egoistic Risk Persp. $[\times 10^3]$ & Altruistic Risk Persp. $[\times 10^3]$& Collective Risk Persp. $[\times 10^3]$\\
        \toprule
        \multicolumn{4}{c}{\texttt{ZAM\_Zip}} \\
        \midrule
            Low & \textbf{4.60} & 15.21 & 6.69 \\
            Moderate & \textbf{7.25} & 12.17 & 8.99 \\
            High  & 11.34 & 12.87 & \textbf{9.68} \\
            \midrule
            \multicolumn{4}{c}{\texttt{ZAM\_Tjunction}} \\
        \midrule
            Low & 5.00 & 3.83 & \textbf{2.50}  \\
            Moderate & 4.71 & 5.72 & \textbf{3.74}\\
            High &  \textbf{4.39} & 7.05 & 4.56 \\
            \midrule
            \multicolumn{4}{c}{\texttt{USA\_US101}} \\
            \midrule
            Low  & \textbf{4.94} & 8.29 & 5.87 \\
            Moderate & \textbf{7.91} & 10.07 & 8.03\\
            High & 12.62 & 12.85 & \textbf{10.96}\\
        \toprule
    \end{tabularx}
\end{table}
\bibliography{lib}
\bibliographystyle{ieeetr}

\end{document}